\documentclass[10pt, a4paper]{article}
\usepackage{lrec2000}
\usepackage{graphicx}
\usepackage[utf8]{inputenc}       %UTF-8 for čšž, also in .bib file!
\usepackage{url}                      %to make links to Web pages

\title{The ParlaSent-BCS dataset of sentiment-annotated parliamentary debates from Bosnia-Herzegovina, Croatia, and Serbia}

\name{Michal Mochtak,$^\ast$ Peter Rupnik,$^\dagger$
  Nikola Ljube\v{s}i\'{c}${}^\dagger {}^\ddagger$}

\address{ $^\ast$Institute of Political Science\\
               University of Luxembourg \\
               2 avenue de l’Université, L-4366 Esch-sur-Alzette \\
               michal.mochtak@uni.lu \\ 
               \\
               $^\dagger$ Department of Knowledge Technologies \\
               Jo\v{z}ef Stefan Institute\\
               Jamova cesta 39, SI-1000 Ljubljana\\
               peter.rupnik@ijs.si \\
               nikola.ljubesic@ijs.si \\
               \\
               $^\ddagger$Faculty of Computer and Information Science\\
               University of Ljubljana \\ 
               Ve\v{c}na pot 113, SI-1000 Ljubljana\\
        }

        \abstract{Expression of sentiment in parliamentary debates is deemed to be significantly different from that on social media or in product reviews. This paper adds to an emerging body of research on parliamentary debates with a dataset of sentences annotated for detection sentiment polarity in political discourse. We sample the sentences for annotation from the proceedings of three Southeast European parliaments: Croatia, Bosnia-Herzegovina, and Serbia. A six-level schema is applied to the data with the aim of training a classification model for the detection of sentiment in parliamentary proceedings. Krippendorff's alpha measuring the inter-annotator agreement ranges from 0.6 for the six-level annotation schema to 0.75 for the three-level schema and 0.83 for the two-level schema. Our initial experiments on the dataset show that transformer models perform significantly better than those using a simpler architecture. Furthermore, regardless of the similarity of the three languages, we observe differences in performance across different languages. Performing parliament-specific training and evaluation shows that the main reason for the differing performance between parliaments seems to be the different complexity of the automatic classification task, which is not observable in annotator performance. Language distance does not seem to play any role neither in annotator nor in automatic classification performance. We release the dataset and the best-performing model under permissive licences.}

\begin{document}

\maketitleabstract

\section{Introduction}

Emotions and sentiment in political discourse are deemed as crucial and influential as substantive policies promoted by the elected representatives~\cite{young2012affective}. Since the golden era of research on propaganda~\cite{lasswell1927,shils1948}, a number of scholars have demonstrated the growing role of emotions on affective polarization in politics with negative consequences for the stability of democratic institutions and the social cohesion~\cite{garrett2014,iyengar2019,mason2015}. With the booming popularity of online media, sentiment analysis has become an indispensable tool for understating the positions of viewers, customers, but also voters~\cite{soler_2012}. It has allowed all sorts of entrepreneurs to know their target audience like never before~\cite{ceron2019}. Experts on political communication argue that the way we receive information and how we process them play an important role in political decision-making, shaping our judgment with strategic consequences both on the level of legislators and the masses~\cite{liu2018}. Emotions and sentiment simply do play an important role in political arenas and politicians have been (ab)using them for decades.

Although there is a general agreement among political scientists that sentiment analysis represents a critical component for understanding political communication in general~\cite{young2012,flores2017,Tumasjan_Sprenger_Sandner_Welpe_2010}, the empirical applications outside the English-speaking world are still rare ~\cite{rauh2018,mohammad2021sentiment}. This is especially the case for studies analyzing political discourse in low-resourced languages, where the lack of out-of-the-box tools creates a huge barrier for social scientists to do such research in the first place~\cite{proksch2019,mochtak2020,rauh2018}. The paper therefore aims to contribute to the stream of applied research on sentiment analysis in political discourse in low-resourced languages. We specifically focus on training classification models for the sentiment analysis in political discourse in the region of Southeast Europe. The paper contributes to an emerging community of datasets of parliamentary debates with a focus on sentence-level sentiment annotation with future downstream applications in mind.

\section{Dataset construction}

\subsection{Background data}

In order to train the classification models, we sampled sentences from three corpora of parliamentary proceedings in the region of former Yugoslavia – Bosnia-Herzegovina~\cite{mochtak_bihcorp2002}\footnote{\url{https://doi.org/10.5281/zenodo.6517697}}, Croatia ~\cite{mochtak_crocorp2022}\footnote{\url{https://doi.org/10.5281/zenodo.6521372}}, and Serbia ~\cite{mochtak_srbcorp2022}\footnote{\url{https://doi.org/10.5281/zenodo.6521648}}. The Bosnian corpus contains speeches collected on the federal level from the official website of the Parliamentary Assembly of Bosnia and Herzegovina~\cite{parlamentarna_skupstina_bih_sjednice_2020}. Both chambers are included – House of Representatives (Predstavnički dom / Zastupnički dom) and House of Peoples (Dom naroda). The corpus covers the period from 1998 to 2018 (2nd – 7th term) and counts 127,713 speeches. The Croatian corpus of parliamentary debates covers debates in the Croatian parliament (Sabor) from 2003 to 2020 (5th – 9th term) and counts 481,508 speeches~\cite{hrvatski_sabor_edoc_2020}. Finally, the Serbian corpus contains 321,103 speeches from the National Assembly of Serbia (Skupština) over the period of 1997 to 2020 (4th – 11th term)~\cite{otvoreni_parlament_pocetna_2020}.

\subsection{Data sampling}

Each speech was processed using the CLASSLA fork of Stanza for processing Croatian and Serbian~\cite{ljubesic-dobrovoljc-2019-neural} in order to extract individual sentences as the basic unit of our analysis. In the next step, we filtered out only sentences presented by actual speakers, excluding moderators of the parliamentary sessions. All sentences were then merged into one meta dataset. As the goal was to sample what can be understood as ``average sentences'', we further subset the sentence meta corpus to only sentences having the number of tokens within the first and third frequency quartile (i.e. being within the interquartile range) of the original corpus ($\sim$3.8M sentences). Having the set of ``average sentences'', we used the Croatian gold standard sentiment lexicon created by~\cite{glavavs2012semi}, translated it to Serbian with a rule-based Croatian-Serbian translator~\cite{klubivcka2016collaborative}, combined both lexicons, and extracted unique entries with a single sentiment affinity, and used them as seed words for sampling sentences for manual annotation. The final pool of seed words contains 381 positive and 239 negative words (neutral words are excluded). These seed words are used for stratified random sampling which gives us 867 sentences with negative seed word(s), 867 sentences with positive seed word(s), and 866 sentences with neither positive nor negative seed words (supposedly having neutral sentiment). We sample 2600 sentences in total for manual annotation. The only strata we use is the size of the original corpora (i.e. number of sentences per corpus). With this we sample 1,388 sentences from the Croatian parliament, 1,059 sentences from the Serbian parliament, and 153 sentences from the Bosnian parliament.

\subsection{Sentence-level data\label{sec:sentence_level_data}}
The focus on sentences as the basic level of the analysis goes against the mainstream research strategies in social sciences which focus either on longer pieces of text (e.g. utterance of ‘speech segment’ or whole documents~\cite{bansal-etal-2008-power,thomas-etal-2006-get}) or coherent messages of shorter nature (e.g. tweets~\cite{Tumasjan_Sprenger_Sandner_Welpe_2010,flores2017}). This approach, however, creates certain limitations when it comes to political debates in national parliaments where speeches range from very short comments counting only a handful of sentences to long monologues having thousands of words. Moreover, as longer text may contain a multitude of sentiments, any annotation attempt must generalize them, introducing a complex coder bias which is embedded in any subsequent analysis. The sentence-centric approach attempts to refocus the attention on individual sentences capturing attitudes, emotions, and sentiment positions and using them as lower-level indices of sentiment polarity in a more complex political narrative. Although sentences cannot capture complex meanings as paragraphs or whole documents do, they usually carry coherent ideas with relevant sentiment affinity. This approach stems from a tradition of content analysis in political science which focuses both on the political messages and their role in political discourse in general ~\cite{burst2022,hutter_politicising_2016,koopmans_political_2006}. 

Unlike most of the literature which approaches sentiment analysis in political discourse as a proxy for position-taking stances or as a scaling indicator~\cite{abercrombie_sentiment_2020,glavas-etal-2017-unsupervised,proksch2019}, a general sentence-level classifier we present in this paper has a more holistic (and narrower) aim. Rather than focusing on a specific policy or issue area, the task is to assign a correct sentiment category to sentence-level data in political discourse with the highest possible accuracy. Only when a good performing model exists, a downstream task can be discussed. We believe it is a much more versatile approach which opens a wide range of possibilities for understanding the context of political concepts as well as their role in political discourse. Furthermore, sentences as lower semantic units can be aggregated to the level of paragraphs or whole documents which is often impossible the other way around (document → sentences). Although sentences as the basic level of analysis are less common in social sciences research when it comes to computational methods \cite{abercrombie_sentiment_2020}, practical applications in other areas exist covering topics such as validation of sentiment dictionaries \cite{rauh2018}, ethos mining \cite{duthie2018}, opinion mining \cite{naderi2016}, or detection of sentiment carrying sentences \cite{onyimadu2013}.

\subsection{Annotation schema\label{sec:anno}}

The annotation schema for labelling sentence-level data was adopted from Batanović et al. \cite{batanovic_versatile_2020} who propose a six-item scale for annotation of sentiment polarity in a short text. The schema was originally developed and applied to SentiComments.SR, a corpus of movie comments in Serbian and is particularly suitable for low-resourced languages. The annotation schema contains six sentiment labels \cite[p.~6]{batanovic_versatile_2020}:

\begin{itemize}
    \item +1	for sentences that are entirely or predominantly positive
    \item –1	for sentences that are entirely or predominantly negative
    \item +M	for sentences that convey an ambiguous sentiment or a mixture of sentiments, but lean 	more towards the positive sentiment in a strict binary classification
    \item –M	for sentences that convey an ambiguous sentiment or a mixture of sentiments, but lean 	more towards the negative sentiment in a strict binary classification
    \item +NS	for sentences that only contain non-sentiment-related statements, but still lean more 	towards the positive sentiment in a strict binary classification
    \item –NS	for sentences that only contain non-sentiment-related statements, but still lean more towards 	the negative sentiment in a strict binary classification

\end{itemize}

Additionally, we also follow the original schema which allowed marking text deemed as sarcastic with a code ``sarcasm''. The benefit of the whole annotation logic is that it was designed with versatility in mind allowing reducing the sentiment label set in subsequent processing if needed. That includes various reductions considering polarity categorization, subjective/objective categorization, change of the number of categories, or sarcasm detection. This is important for various empirical tests we perform in the following sections.

\subsection{Data annotation\label{sec:dataanno}}

Data were annotated in two waves, with 1300 instances being annotated in each. Annotation was done via a custom online app. The first batch of 1300 sentences was annotated by two annotators, both being native speakers of Croatian, while the second batch was annotated only by one of them. The inter-coder agreement measured using Krippendorff's alpha in the first round was 0.599 for full six-item annotation scheme, 0.745 for the three-item annotation schema (positive/negative/neutral), and 0.829 for the two-item annotation schema focused on the detection of only negative sentiment (negative/other). The particular focus on negative sentiment in the test setting is inspired by a stream of research in political communication which argues that negative emotions appear to be particularly prominent in the context of forming the human psyche and its role in politics \cite{young2012}. More specifically, political psychologists have found that negative political information has a more profound effect on attitudes than positive information as it is easier to recall and is more useful in heuristic cognitive processing for simpler tasks \cite{baumeister_bad_2001,utych_negative_2018}.

Before the second annotator moved to annotate the second batch of instances, hard disagreements, i.e. disagreements pointing at a different three-class sentiment, where +NS and -NS are considered neutral, were resolved together by both annotators through a reconciliation procedure.

\subsection{Dataset encoding}

The final dataset, available through the CLARIN.SI repository, contains the following metadata:

\begin{itemize}
    \item \texttt{sentence} that is annotated
    \item \texttt{country} of origin of the sentence
    \item annotation \texttt{round} (first, second)
    \item annotation of \texttt{annotator1} with one of the labels from the annotation schema presented in Section~\ref{sec:anno}
    \item annotation of \texttt{annotator2} following the same annotation schema
    \item annotation given during \texttt{reconciliation} of hard disagreements
    \item the three-way \texttt{label} (positive, negative, neutral) where +NS and -NS labels are mapped to the neutral class
    \item the \texttt{document\_id} the sentence comes from
    \item the \texttt{sentence\_id} of the sentence
    \item the \texttt{date} the speech was given
    \item the \texttt{name}, \texttt{party},  \texttt{gender}, \texttt{birth\_year} of the speaker
    \item the \texttt{split} (train, dev, or test) the instance has been assigned to (described in more detail in Section \ref{sec:split}
\end{itemize}

%The document and sentence identifiers will be useful 

The final dataset is organized in a JSONL format (each line in the file being a JSON entry) and is available under the CC-BY-SA 4.0 license.

%* Input: Michal's xls: 2 sheets. Sheet 1 has 2 annotations and gold label, Sheet 2 only one annotation
%* Labels: 'soft disagreement' 'Negative' 'Positive' 'hard disagreement' 'N Neutral' 'P Neutral' 'M Negative' 'M Positive'
%* Labels in sheet 1 were overwritten with values from `reconciliation hard` column where available
%* Labels were downsampled to 3: Negative, Positive, Neutral
%* Where the two annotators were in soft disagreement, the downsampled labels were equal (e.g.: one label was `M Neutral', the other was `Neutral'. After downsampling, the disagreement vanishes)
%* Composition: sheet 1: Negative    692, Neutral     410, Positive    198. Sheet 2: Negative    666, Neutral     362, Positive    272
%* Splitting: test (300 instances) and dev (150 instances) were acquired from Sheet 1, stratified by country of origin. The rest was allocated to train.
%* Results were inspected for consistency between splits on country, party, and sentiment. 

\section{Experiments}

\subsection{Data splits\label{sec:split}}

For performing current and future experiments, the dataset was split into the train, development and test subsets. The development subset consists of 150 instances, while the test subset consists of 300 instances, both using instances from the first annotation round, where two annotations per instance and hard disagreement reconciliations are available. The training data consists of the remainder of the data from the first annotation round and all instances from the second annotation round, summing to 2,150 instances.

While splitting the data, stratification was performed on the variables of three-way sentiment, country, and party. With this we can be reasonably sure that no specific strong bias regarding sentiment, country or political party is present in any of the three subsets.

\subsection{Experimental setup}

In our experiments we investigate the following questions: (1) how well can different technologies learn our three-way classification task, (2) what is the difference in performance depending on which parliament the model is trained or tested on, and (3) is the annotation quality of the best performing technology high enough to be useful for data enrichment and analysis.

We investigate our first question by comparing the results on the following classifiers: fastText~\cite{joulin2016bag} with pre-trained CLARIN.SI word embeddings~\cite{11356/1205}, the multilingual transformer model XLM-Roberta~\cite{xlmroberta},\footnote{\url{https://huggingface.co/xlm-roberta-base}} the transformer model pre-trained on Croatian, Slovenian and English cseBERT~\cite{ulcar-robnik2020finest},\footnote{\url{https://huggingface.co/EMBEDDIA/crosloengual-bert}}, and the transformer model pre-trained on Croatian, Bosnian, Montenegrin and Serbian BERTić~\cite{ljubesic-lauc-2021-bertic}.\footnote{\url{https://huggingface.co/classla/bcms-bertic}}

While comparing the different classification techniques, each model was optimized for the epoch number hyperparameter on the development data, while all other hyperparameters were kept default. For training transformers, the simpletransformers library\footnote{\url{https://simpletransformers.ai}} was used.

The second question on parliament specificity we answer by training separate models on Croatian sentences only and Serbian sentences only, evaluating each model both on Croatian and on Serbian test sentences. We further evaluate the model trained on all training instances separately on instances coming from each of the three parliaments.

For our third question on the usefulness of the model for data analysis, we report confusion matrices, to inform potential downstream users of the model's per-category performance.

\section{Results}

\subsection{Classifier comparison\label{sec:comp}}

We report the results of our text classification technology comparison in Table~\ref{tab:main}. The results show that transformer models are by far more capable than the fasttext technology relying on static embeddings only. Of the three transformer models, the multilingual XLM-RoBERTa model shows to have a large gap in performance to the two best-performing models. Comparing the cseBERT and the BERTić model, the latter manages to come on top with a moderate improvement of 1.5 points in macro-F1. The difference in the results of the two models is statistically significant regarding the MannWhitney U test~\cite{mann1947test}, with a p-value of 0.0053.

\begin{table}
\begin{center}
\begin{tabular}{|l|l|}
\hline
model                      &       macro F1    \\
\hline
\textbf{classla/bcms-bertic}        &  $\textbf{0.7941 ± 0.0101}^{**}$ \\
EMBEDDIA/crosloengual-bert &  0.7709 ± 0.0113 \\
xlm-roberta-base           &  0.7184 ± 0.0139\\
fasttext + CLARIN.SI embeddings            & 0.6312 ± 0.0043 \\
\hline
 
\end{tabular}

\end{center}
\caption{\label{tab:main}Results of the comparison of various text classification technologies. We report macro-F1 mean and standard deviation over 6 runs with the model-specific optimal number of training epochs. The distributions of results of the two best performing models are compared with the Mann-Whitney U test (** $p<0.01$).}
\end{table}

\subsection{Parliament dependence}

We next investigate the dependence of the results on from which parliament the training and the testing data came. Our initial assumption was that the results are dependent on whether the training and the testing data come from the same or a different parliament, with same-parliament results being higher. We also investigate how the model trained on all data performs on parliament-specific test data.

\subsubsection{Impact of training data}

We perform this analysis on all three transformer models from Section~\ref{sec:comp}, hoping to obtain a deeper understanding of parliament dependence on our task. We train and test on data from the Croatian and the Serbian parliament only as the Bosnian parliament's data are not large enough to enable model training.

In Table~\ref{tab:cross} we report the results grouped by model and training and testing parliament. To our surprise, the strongest factor shows not to be whether the training and testing data come from the same parliament, but what testing data are used, regardless of the training data. This trend is to be observed regardless of the model used.

The results show that Serbian test data seem to be harder to classify, regardless of what training data are used, with a difference of $\sim$9 points in macro-F1 for the BERTić model. The difference is smaller for the cseBERT model, $\sim$7 points, but the results on any combination of training and testing data are also lower than those obtained on the BERTić model.

We have additionally explored the possibility of a complexity bias of Serbian test data in comparison to Serbian training data by performing different data splits, but the results obtained were very similar to those presented here. Serbian data seem to be harder to classify in general, which is observed when performing inference over Serbian data. Training over Serbian data still results in a model comparably strong to that based on Croatian training data. Important to note is that the Croatian data subset is 30\% larger than the Serbian one.

To test whether the Serbian data complexity goes back to challenges during data annotation, or whether it is rather the models that struggle with inference over Serbian data, we calculated the Krippendorff IAA on data from each parliament separately. The agreement calculation over the ternary classification schema resulted in an IAA for Bosnian data of 0.69, Croatian data of 0.733, and Serbian data of 0.77. This insight proved that annotators themselves did not struggle with Serbian data as these had the highest IAA. The complexity that can be observed in the evaluation is obviously due to classifiers struggling with Serbian data, something we cannot explain at this point, but that will have to be taken under consideration in future work.

\begin{table}
\centering
\begin{tabular}{|c|cc|}
\hline
\multicolumn{3}{|c|}{XLM-RoBERTa}\\
\hline
train \textbackslash test & HR & SR \\
\hline
HR &  0.7296 ± 0.0251 &  0.6128 ± 0.0341 \\
SR &  0.7323 ± 0.0282 &  0.6487 ± 0.0203 \\
\hline
\multicolumn{3}{|c|}{cseBERT}\\
\hline
train \textbackslash test & HR & SR \\
\hline
HR &  0.7748 ± 0.0174 &  0.7146 ± 0.0175 \\
SR &  0.7762 ± 0.0114 & 0.6989 ± 0.0275  \\
\hline
\multicolumn{3}{|c|}{BERTić}\\
\hline
train \textbackslash test & HR & SR \\
\hline
HR & 0.8147 ± 0.0083 &  0.7249 ± 0.0105 \\
SR &  0.7953 ± 0.0207 &  0.7130 ± 0.0278 \\
\hline
\end{tabular}
\caption{\label{tab:cross}Comparison of the three transformer models when trained and tested on data from the Croatian or Serbian parliament. Average macro-F1 and standard deviation over 6 runs is reported.}
\end{table}

%Rows are training parliament, columns are test parliaments
%
%\begin{tabular}
%& HR & SR \\
%HR & \\
%SR & \\
%\end{tabular}

%\subsubsection{Fasttext}
%
%General experiment results: train on train split, eval on test: 0.4715 +/- 0.0090 macro F1. Sample size: 50.

%\begin{tabular}{|c|cc|}
%\hline
%train \textbackslash test & HR & SR \\
%\hline
%HR & 0.4159 $\pm$ 0.0093 & 0.4042 $\pm$ 0.0230\\
%SR & 0.3121 $\pm$ 0.0101& 0.2617 $\pm$ 0.0158\\
%\hline
%\end{tabular}
%
%Sample size: 50.
%
%
%\subsubsection{Fasttext with embeddings}
%
%General experiment results: train on train split, eval on test: 0.6312 +/- 0.0043 macro F1. Sample size: 10.
%
%\begin{tabular}{|c|cc|}
%\hline
%train \textbackslash test & HR & SR \\
%\hline
%HR & 0.6379 $\pm$ 0.0097 & 0.6356 $\pm$ 0.0073\\
%SR & 0.6194 $\pm$ 0.0029 & 0.5781 $\pm$ 0.0098\\
%\hline
%\end{tabular}

%Sample size: 5. (Models with embeddings train longer).

\subsubsection{Impact of testing data}

In the next set of experiments, we compare the performance of BERTić classifiers trained over all training data, but evaluated on all and per-parliament testing data. Beyond this, we train models over the ternary schema that we have used until now (positive vs. neutral vs. negative), but also the binary schema (negative vs. rest), given our special interest in identifying negative sentences, as already discussed in Section~\ref{sec:dataanno}

We report results on test data from each of the three parliaments, including the Bosnian one, which, however, contains only 18 testing instances, so these results have to be taken with caution.

The results presented in Table~\ref{tab:ternary_binary} show again that the Serbian data seem to be the hardest to classify even when all training data are used. Bosnian results are somewhat close to the Serbian ones, but caution is required here due to the very small test set. This level of necessary caution regarding Bosnian test data is also visible from the five times higher standard deviation in comparison to the results of the two other parliaments. Croatian data seem to be easiest to classify, with an absolute difference of 9 points between the performance on Serbian and Croatian test data. Regarding the binary classification results, these are, as expected, higher than those of the ternary classification schema with an macro-F1 of 0.9 when all data are used. The relationship between specific parliaments is very similar to that observed using the ternary schema.

\begin{table}
\centering
\begin{tabular}{|c|c|c|}
\hline
test & ternary & binary \\
\hline
all & 0.7941 ± 0.0101 & 0.8999 ± 0.0120 \\
HR & 0.8260 ± 0.0186 & 0.9221 ± 0.0153 \\
BS & 0.7578 ± 0.0679 & 0.9071 ± 0.0525\\
SR & 0.7385 ± 0.0170 & 0.8660 ± 0.0150\\
\hline
\end{tabular}
\caption{\label{tab:ternary_binary} Average macro-F1 and standard deviation of 6 runs of the BERTić model, trained on all training data, and evaluated on varying testing data.}
\end{table}

\subsection{Per-category analysis}

\begin{figure}
\begin{center}
    \includegraphics[width=0.8\columnwidth]{ 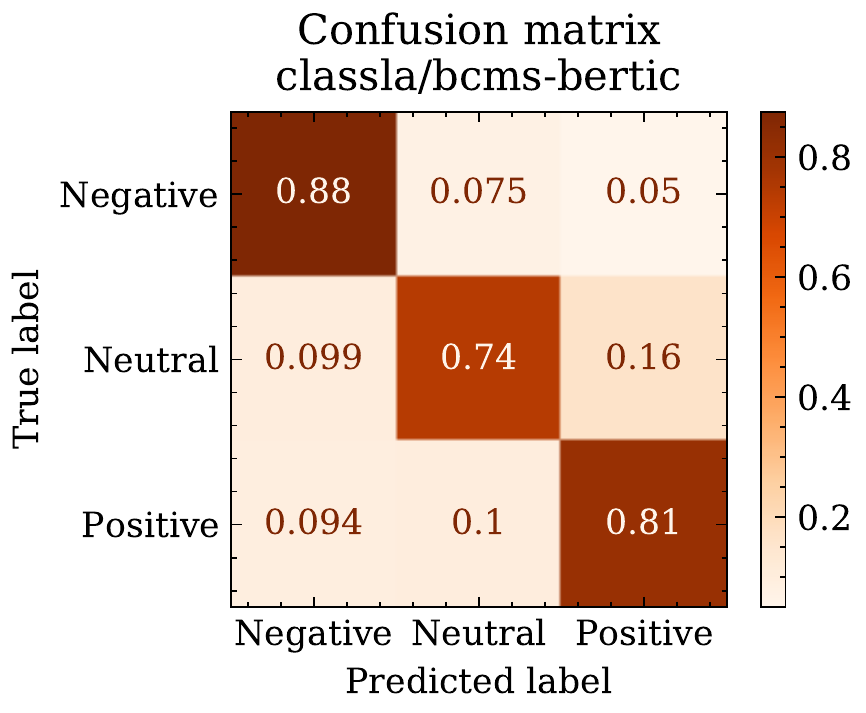}
    \includegraphics[width=0.8\columnwidth]{ 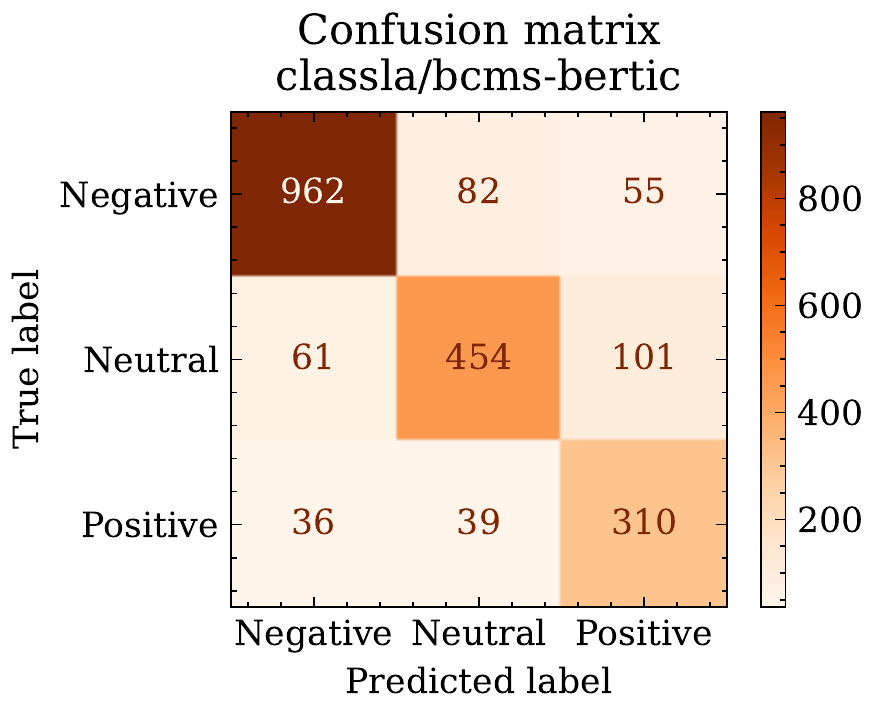}
\end{center}
\caption{\label{fig:confusion}Row-normalised and raw-count confusion matrix of the BERTić results on the ternary schema}
\end{figure}

Our final set of experiments investigates the per-category performance both on the ternary and the binary classification schema. We present the confusion matrices on the ternary schema, one row-normalized, another with raw counts, in Figure~\ref{fig:confusion}. As anticipated, the classifier works best on the negative class, with 88\% of negative instances properly classified as negative. Second by performance is the positive class with 81\% of positive instances being labelled like that, while among the neutral instances 3 out of 4 instances are correctly classified. Most of the confusion between classes occurs, as expected, between the neutral and either of the two remaining classes.

\begin{figure}
\begin{center}
    \includegraphics[width=0.8\columnwidth]{ 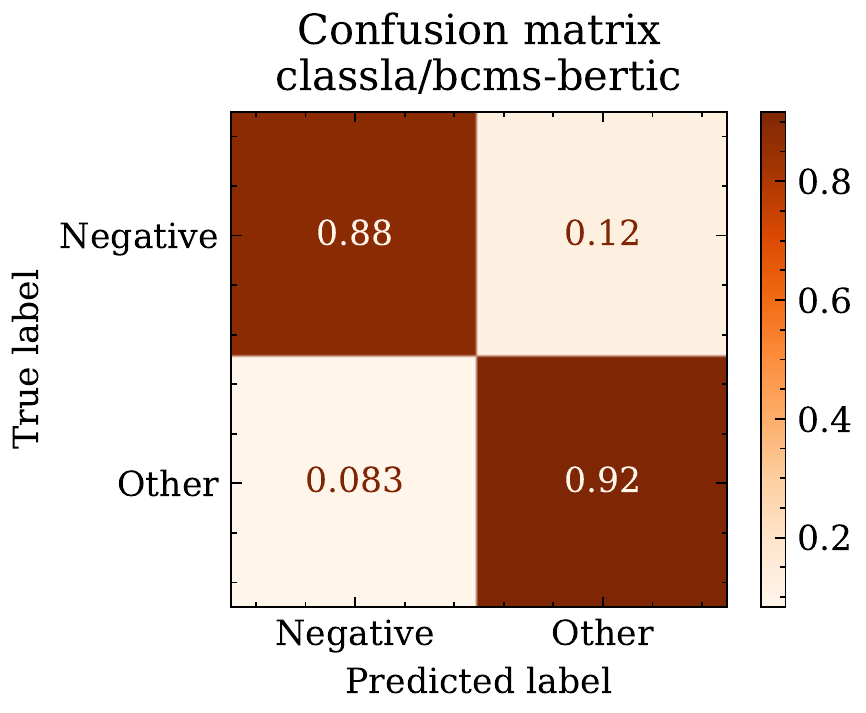}   
    \includegraphics[width=0.8\columnwidth]{ 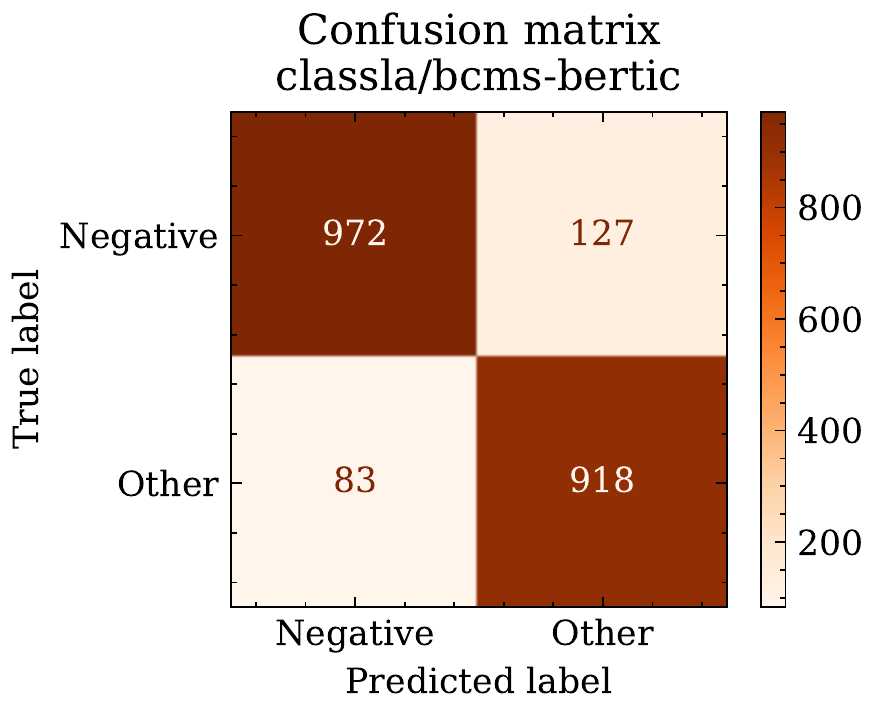} 
\end{center}
\caption{\label{fig:confusion_binary}Row-normalised and raw-count confusion matrix of the BERTić results on the binary schema}
\end{figure}

The binary confusion matrices, presented in Figure~\ref{fig:confusion_binary} show for a rather balanced performance on both categories. On each of the categories recall is around 0.9, with a similar precision given the symmetry of the confusions.

When comparing the output of the ternary and the binary model, the ternary model output mapped to a binary schema performs slightly worse than the binary model, meaning that practitioners should apply the binary model if they are interested just in distinguishing between negative and other sentences.

Although any direct comparisons are hard to make, the few existing studies which performed text classification on sentence-level data, report much worse results. \newcite{rauh2018} found that when three annotators and three sentiment dictionaries were compared on a ternary classification task (positive/negative/neutral), they agreed only in one-quarter of the 1,500 sentences. Using heuristic classifiers based on the use of statistical and syntactic clues, \newcite{onyimadu2013} found that on average, only 43\% of the sentences were correctly annotated for their sentiment affinity. The results of our experiments are therefore certainly promising. Especially when it comes to the classification of negative sentences, the model has 1 in 10 sentence error rate which is almost on par with the quality of annotation performed by human coders.

\section{Conclusion}

This paper presents a sentence-level dataset of parliamentary proceedings, manually annotated for sentiment via a six-level schema. The good inter-annotator agreement is reported, and the first results on the automation of the task are very promising, with n macro-F1 of $\sim$0.8 on the ternary schema and $\sim$0.9 on the binary schema. The difference in performance across the three parliaments is observed, but visible only during inference, Serbian data being harder to make predictions on, while for modelling, all parliaments seem to be similarly useful. One limitation of our work is the following: our testing data have been sampled as the whole dataset, with a bias towards mid-length sentences, and sentences containing sentiment words. Future work should consider preparing a sample of random sentences, or,  even better, consecutive sentences, so that the potential issue of lack of a wider context is successfully mitigated as well.

In general, the reported results have several promising implications for applied research in political science. First of all, it allows a more fine-grained analysis of political concepts and their context. A good example is a combination of the KWIC approach with sentiment analysis with a focus on examining the tone of a message in political discourse. This is interesting for both qualitatively and quantitatively oriented scholars. Especially the possibility of extracting numeric assessment of the classification model (e.g. class probability) is particularly promising for all sorts of hypothesis-testing statistical models. Moreover, sentence-level analysis can be combined with the findings of various information and discourse theories for studying political discourse focused on rhetoric and narratives (e.g. beginning and end of a speech are more relevant than what comes in the middle). Apart from concept-driven analysis, the classification model can be used for various research problems ranging from policy position-taking to ideology detection or general scaling tasks \cite{abercrombie-batista-navarro-2020-parlvote,glavas-etal-2017-unsupervised,proksch2019}. Although each of these tasks requires proper testing, the performance of the trained models for such applications is undoubtedly promising. 

As a part of our future work, we plan to test the usefulness of the predictions on a set of downstream tasks. The goal is to analyze the data from all three parliaments (Bosnia-Herzegovina, Croatia, and Serbia) in a series of tests focused on replication of the results from the existing research using mostly English data. Given the results we obtained, we aim to continue our research using the setup with the model trained on cross-country data. Furthermore, the three corpora we have used in this paper will be extended as a part of ParlaMint II project.

We make the ternary and binary BERTić models trained on all available training available via the HuggingFace repository\footnote{\url{https://huggingface.co/classla/bcms-bertic-parlasent-bcs-ter}}\footnote{\url{https://huggingface.co/classla/bcms-bertic-parlasent-bcs-bi}} and make the dataset available through the CLARIN.SI repository.\footnote{\url{http://hdl.handle.net/11356/1585}}

\bibliographystyle{lrec2000}
\bibliography{xample-en} 

\end{document}